\documentclass{article} 
\pdfoutput=1
\usepackage{nips13submit_e,times}
\usepackage{hyperref}
\usepackage{url}
\usepackage{comment}
\usepackage{slashbox}

\title{Analyzing noise in autoencoders and deep networks}

\author{
Ben Poole$^*$,\; Jascha Sohl-Dickstein$^\dagger$, \;Surya Ganguli$^\dagger$\\
Departments of Computer Science$^*$ and Applied Physics$^\dagger$ \\
Stanford University, Stanford, CA 94305\\
\texttt{poole@cs.stanford.edu} ,\texttt{ \{jascha, sganguli\}@stanford.edu}\\
}

\nipsfinalcopy 

\usepackage{amsmath}
\usepackage{amssymb}
\usepackage{graphicx}

\let\oldmarginpar\marginpar
\renewcommand\marginpar[1]{\-\oldmarginpar[\raggedleft\footnotesize #1]%
{\raggedright\footnotesize #1}}

\newcommand{\ex}[1] {\mathbb{E}\left[#1\right]}
\newcommand{\exx}[2] {\mathbb{E}_{#2}\left[#1\right]}

\newcommand{\var}[1] {\mathrm{Var}(#1)}
\newcommand{\tr}[1] {\mathrm{tr}(#1)}
\newcommand{\diag}[1] {\mathrm{diag}(#1)}
\newcommand{\diags}[1] {\mathrm{diag}(#1))}

\newcommand{\given} {\,|\,}

\newcommand{\eps}{\epsilon}

\renewcommand{\xi}{x^{(i)}}
\newcommand{\hti}{\tilde{h}^{(i)}}
\newcommand{\xti}{\tilde{x}^{(i)}}
\newcommand{\hhi}{{h}^{(i)}}

\newcommand{\pdot}{\odot}

\newcommand{\ei}{\epsilon_\mathcal{I}}
\newcommand{\ez}{\epsilon_\mathcal{Z}}
\newcommand{\eh}{\epsilon_\mathcal{H}}
\newcommand{\ci}{{c_\mathcal{I}}}
\newcommand{\cz}{{c_\mathcal{Z}}}
\newcommand{\ch}{{c_\mathcal{H}}}
\newcommand{\x}{\odot}
\newcommand{\tih}{\tilde{h}}
\newcommand{\tix}{\tilde{x}}
\newcommand{\tiz}{\tilde{z}}
\begin{document}

\maketitle

\begin{abstract}
Autoencoders have emerged as a useful framework for unsupervised learning of internal representations, and a wide variety of apparently conceptually disparate regularization techniques have been proposed to generate useful features.  Here we extend existing denoising autoencoders to additionally inject noise before the nonlinearity, and at the hidden unit activations. We show that a wide variety of previous methods, including denoising, contractive, and sparse autoencoders, as well as dropout can be interpreted using this framework. This noise injection framework reaps practical benefits by providing a unified strategy to develop new internal representations by designing the nature of the injected noise.  We show that noisy autoencoders outperform denoising autoencoders at the very task of denoising, and are competitive with other single-layer techniques on MNIST, and CIFAR-10. We also show that types of noise other than dropout improve performance in a deep network through sparsifying, decorrelating, and spreading information across representations. 


\end{abstract} 
\section{Introduction}
Regularization through noise \cite{Bishop:1995tf, An:1996wk} has regained focus recently in the training of supervised neural networks. Randomly dropping out units while performing backpropagation has been shown to consistently improve the performance of large neural networks  \cite{Anonymous:H68d5mS5,Hinton:2012tv}. 
Stochastic pooling, where a set of input units are gated based off their activations, has also been shown to improve performance in convolutional nets over noiseless max and average pooling \cite{Zeiler:2013tna, Lee:2009tm}. The role of input noise in training unsupervised networks has also been extensively explored in recent years \cite{Vincent:2010vu}. Injecting noise into the input layer of autoencoders has been shown to yield useful representations in these denoising autoencoders.

Motivated by the success of noise injection at the input layer in autoencoders, and at the hidden layers in supervised learning settings, we systematically explore the role of noise injection at all layers in unsupervised feature learning models. As we shall see, this provides a unified framework for unsupervised learning based on the principle that hidden representations should be robust to noise. This yields an extension of prior methods for regularizing autoencoders that we call the noisy autoencoder (NAE). For certain types of NAEs, we are able to marginalize out the noise, and derive a set of penalties that relate noise injection with contractive autoencoders, sparse autoencoders, dropout, and ICA. Experiments on MNIST and CIFAR-10 validate the effectiveness of noisy autoencoders at learning useful features for classification.

Building upon the recent success of dropout, we also experiment with further supervised fine-tuning of NAEs in which the noise is also injected at the supervised stage. We show that training noisy autoencoders with dropout noise, and supervised fine-tuning with dropout noise, allows us to waste less capacity in larger autoencoder networks. We also show that purely supervised training with additive Gaussian noise beats dropout on MNIST. We compare the effect of these different types of noise, and argue that these results point to an interaction between noise in unsupervised and supervised learning that may be more complex. These results suggest that we may be able to optimize stacked learning strategies by introducing different types of noise for the unsupervised pretraining relative to the supervised fine-tuning.




\section{Autoencoder Framework}
\subsection{Autoencoders and Denoising Autoencoders}
An autoencoder is a type of one layer neural network that is trained to reconstruct its inputs. In the complete case, this can trivially be accomplished using an identity transformation. However, if the network is constrained in some manner, then the autoencoder tends to learn a more interesting representation of the input data that can be useful in other tasks such as object recognition. The autoencoder consists of an encoder that maps inputs to a hidden representation: $f(x) = s_f(Wx+b)$, and a decoder that maps the hidden representation back to the inputs: $g(h) = s_g(W'h + d)$. The composition of the encoder and decoder yield the reconstruction function: $r(x) = g(f(x))$.
The typical training criterion for autoencoders is minimizing the reconstruction error, $\sum_{x\in\mathcal{X}}L(x, r(x))$ with respect to some loss $L$, typically either squared error or the binary cross-entropy \cite{Bengio:2013ux}.

Denoising autoencoders (DAEs) are an extension of autoencoders trained to reconstruct a clean version of an input from its corrupted version \cite{Vincent:2010vu}. The denoising task requires the network to learn representations that are able to remove the corrupted noise. Prior work has shown that DAEs can be stacked to learn more complex representations that are useful for a variety of tasks \cite{Cho:2013wc,Vincent:2010vu,Chen:2012tq}. The objective for the DAE can be written as $\sum_{x\in\mathcal{X}} \ex{L(x, r(\tilde{x}))}$ where $\tilde{x}$ is a corrupted version of the input $x$.




\subsection{Noisy Autoencoder Model}

\begin{figure}[h]
\begin{center}
\includegraphics[width=0.8\textwidth]{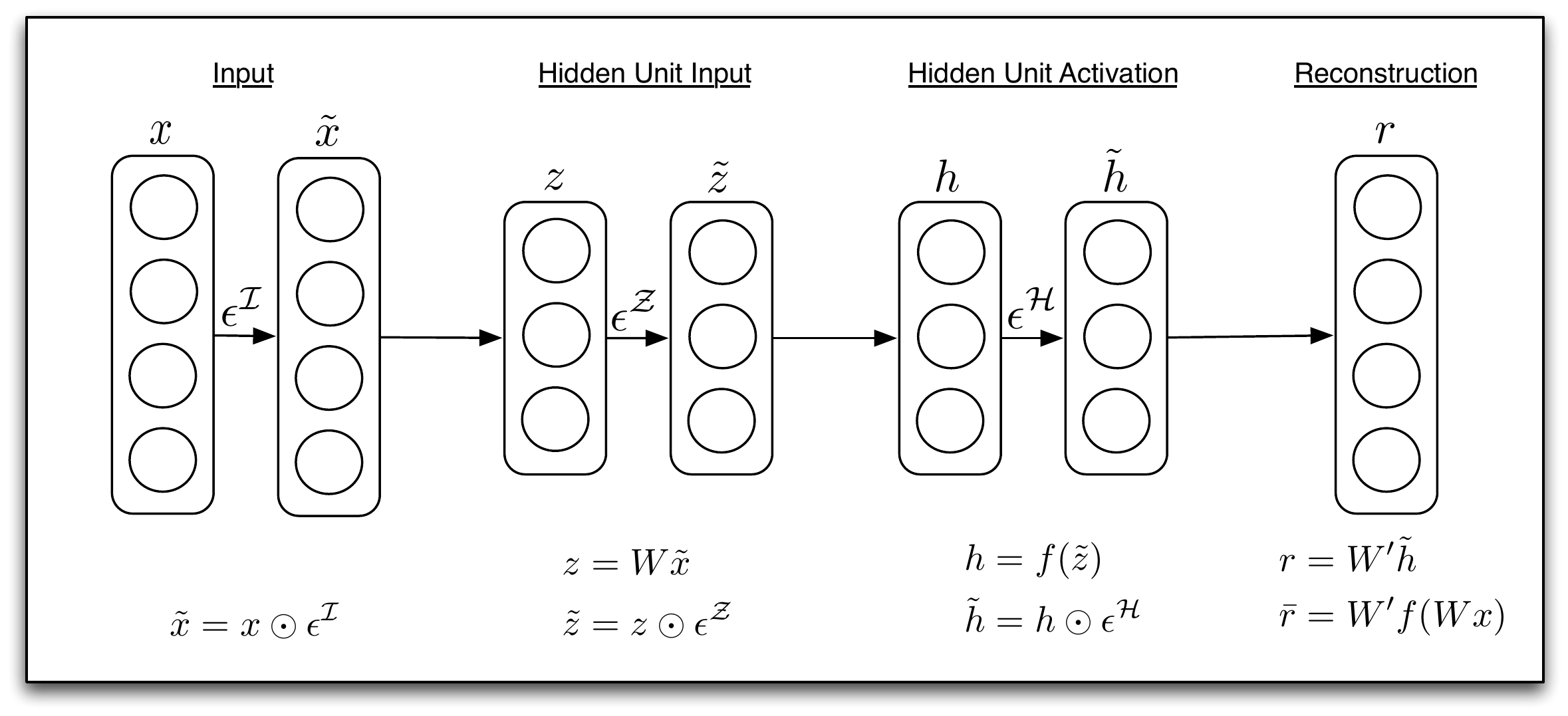}
\end{center}
\caption{Noisy autoencoder (NAE) model. Tildes indicate corrupted versions of the previous layer. }
\label{fig:model}
\end{figure}

Inspired by the recent work on dropout, we extend denoising autoencoders to allow for the injection of additional noise at the input and output of the hidden units. We call these models noisy autoencoders (NAEs) as their hidden representations are stochastic, and no longer a deterministic function of the input. Injecting noise into both the inputs and hidden representations of autoencoders has been proposed for linear networks in prior work by \cite{Inayoshi:2005iz}, but has not been analyzed in detail for nonlinear representations. We parameterize the noise in the NAE as a tuple $(\ei,\eh,\ez)$ that characterizes the distribution of the noises corrupting the  input, hidden unit inputs, and hidden activations respectively (see Figure \ref{fig:model}). We define the encoder and reconstruction function for the NAE model as: 
\begin{align}
\tilde{h}(x,\ei,\ez,\eh) &=  \tilde{f}_\theta(x) = s_f\left(\left(W(x\x \ei)+b\right)\x\ez\right)\x \eh\\
r(x,\ei,\ez,\eh) &= s_g\left(\left[W'\tilde{h}(x,\ei,\ez,\eh) + d\right]\pdot \eps_H \right) 
\end{align}
where $\x$ denotes either addition or multiplication, $s_f$ and $s_g$ are elementwise nonlinearities and we use tildes to denote corrupted versions of variables. As with the DAE, we optimize the expected reconstruction error when training: $\sum_{x\in\mathcal{X}}\ex{L(x,\tilde{r}(x,\ei,\ez,\eh)}$.

When using the NAE to extract features or perform denoising on testing data we can compute the expectation of the noisy hidden activation or reconstruction by sampling from the NAE. However, this can be prohibitively costly on large datasets. Instead, we can approximate the expectation by scaling each of the corrupted variables by their expectation as in dropout. In practice we only use noise where the corruption of the input variable does not alter the mean so that no scaling is needed. The test-time hidden representation and reconstruction are then computed in the same way as the vanilla autoencoder.




\subsection{Relationships between noise types}

Due to the parameterization of noise in the noisy autoencoder, there are many possible choices of noise $(\ei, \ez, \eh)$, that will yield equivalent effective noise on the reconstruction. In particular, we can always rewrite a NAE so that the only source of noise is on the hidden activations $\eh$.

To analyze the effect of introducing noise before the encoder nonlinearity, we perform a first-order Taylor expansion of the encoder function:
$s_f(W\tilde{x} + \ez) \approx s_f(W\tilde{x}) +\diags{ s_f'(W\tilde{x})} \ez$. 
Thus, for small noise, adding noise $\ez \sim Q$ to the linear encoder part is equivalent to having hidden noise: $\eh \sim \diags{ s_f'(W\tilde{x})} Q$. 

If the input noise is Gaussian with some covariance $\Sigma$, then the equivalent hidden unit input noise is Gaussian with covariance $W\Sigma W^T$. If the singular values of $W$ are sufficiently small, then we can use the above result to get the effective hidden noise: $\eh \sim N(0, \diag{s_f'(Wx)}W\Sigma W^T)$. If the input noise is dropout (multiplicative bernoulli), then we can use the result from \cite{Wang:0xFoHXs6} to approximate $\ez$ as a Gaussian with some mean and diagonal covariance. These simple relationships allow us to identify the covariance structure of noise at the hidden unit activations that corresponds to input and hidden unit input noise.


\subsection{Marginalized Noise Penalties}
To understand the impact that these additional sources of noise have on autoencoder training we analyze a simplified NAE. We assume that the decoder is linear ($s_g(x)=x$), the loss function used is squared error, and that all the corruptions are applied independently to each neuron. This allows us to exactly marginalize out the noise on the hidden unit activations using a result from \cite{Bishop:1995tf}:
\begin{equation}
\ex{ \| x - r(x,\eh)\|^2} = \| x- r(x)\|^2 + \tr{WW^T\var{\tilde{h}|h}}
\end{equation}

We can then apply the approximation results from above to yield cost functions corresponding to marginalizing out each different type of noise independently: 
\begin{align}
L(W) &\approx\sum_{x\in\mathcal{X}}  \left[\bar{r}(x) + \ch(x)  + \cz(x) + \ci(x)\right]\\
\ch(x) &= \textstyle\sum_{i=1}^d \var{\tih_i|h} \|w_i\|^2 \\
\cz(x) &\approx  \textstyle\sum_{i=1}^d \var{\tiz_i|z} (f'(w_i^Tx) \|w_i\|)^2\\
\ci(x) &\approx \textstyle\|WW^T\diag{f'(Wx)}\var{\tix|x}\|_F^2
\end{align}
where $d$ is the number of hidden units, and $w_i'$ is the $i$th column of $W'$. Note that the penalty $\ch(x)$ is exact, while the penalties for noise at the input and hidden unit inputs will only be accurate when the variance of their noises are small. These penalties allow us to relate injection of noise in a NAE with regularization penalties from other regularized autoencoders.



\section{Connections to prior work}
The noisy autoencoder and associated marginalized noise penalties provide a framework for comparing many types of regularized autoencoders, and help to explain the utility of injecting noise into more general neural networks. 

\subsection{Regularized Autoencoders}
The marginalized input noise penalty from the noisy autoencoder provides an intuition in the success of {\bf denoising autoencoders}. If we use tied weights, and additive Gaussian corrupting noise with variance $\sigma^2$ then the penalty becomes:
\begin{equation}
\ci(x) =  \|WW^T\diag{s_f'(Wx)}\sigma^2\|_F^2=\sigma^4\textstyle\sum_{i,j} \left(\|w_i^Tw_j\| (s_f'(w_i^Tx))\right)^2
\end{equation}
This penalty encourages the hidden units to learn orthogonal representations of the input, and provides a contractive-like penalty on individual filters. A similar type of penalty is found when learning overcomplete representations with reconstruction ICA, where they use $\|WW^T-I\|_F^2$ to encourage a diverse set of filters \cite{Anonymous:DmtqWbpa}.

When stacking denoising autoencoders, we end up with a similar structure to a noisy autoencoder. In the case of building a two layer network, the standard practice is to first train a denoising autoencoder on the input, then compute the first layer representation by encoding the clean inputs. Training a denoising autoencoder on the second layer will corrupt the first layer representation, but will not impact the prior encoding model learned for the first layer. In contrast, training a NAE with noise on the hidden unit activations allows for the first layer representation to learn to be robust to noise in its hidden representation.

{\bf Contractive autoencoders} aim to learn a representation that is insensitive to small changes in the input space \cite{Anonymous:g1ChF3mx}. They penalize the Frobenius norm of the Jacobian of the encoder function, $\lambda ||J_f(x)||_F^2=\sum_{i=1}^d  s_f'(w_i^Tx) \|w_i\|^2$.
If we inject additive white Gaussian noise at the hidden inputs, then we recover the same penalty. Alternatively, we can inject additive noise with covariance $\diag{s_f'(Wx)^2}$ at the hidden unit activations and recover the exact penalty. This result has been previously reported in \cite{Rifai:2011td}, and motivated the contractive penalty.

{\bf Sparse autoencoders} force all hidden units to have similar mean activations\cite{Goodfellow:2009vda} . We cannot directly relate this penalty to a form of noise, but we can recover a penalty that encourages sparsity on hidden unit activations. If we inject additive Gaussian noise on the activations of the hidden units with variance equal to the uncorrupted hidden unit activation then the marginalized noise penalty becomes: $\ch(x) =\sum_{i=1}^d h_i \|w_i\|^2$. If activations are non-negative then this penalty will force many of the hidden unit activations to zero. We note that experimental results from neuroscience have shown that cortical neurons exhibit Poisson-like noise statistics with a Fano factor of 1. 



\subsection{Dropout}
Dropout is a simple regularizer used in training neural networks that applies multiplicative bernoulli noise to all units in a neural network \cite{Hinton:2012tv}. This noise has been shown to effectively regularize deep neural networks, and has been linked to preventing co-adaptation of neurons and model averaging \cite{Anonymous:5Kr-Pl7w}. The primary motivation for NAEs was the success of dropout in improving generalization performance. We can analyze the effect of dropout noise in NAEs by computing the corresponding marginalized noise penalty: 
$\ch(x) =  \sum_{i=1}^d \var{\tih_i|h} \|w_i'\|^2 = p(1-p)\sum_{i=1}^d (h_i\|w_i'\|)^2$
Thus dropout in a NAE shrinks the average size of the projective fields ($h_i\|w_i'\|$) of the hidden units. Shrinking the size of the projective field helps to reduce the sensitivity of the reconstruction to dropping out hidden units.

\subsection{Other models}
Recent work in neuroscience has shown that single-layer models trained with input and output noise and optimized to maximize mutual information yield receptive fields resembling those found in biological systems \cite{Doi:2012vi,Karklin:N4TxuNDQ}. Similar to our work, these models show the importance of input noise and hidden activation noise on learning representations. 

In {\bf semantic hashing}, Gaussian noise is injected at an intermediate layer to learn binary representations \cite{Salakhutdinov:2007wy}. These representations allow for fast matching in lower-dimensional space using binary codes. To the best of our knowledge the advantage of adding noise in terms of accuracy of the fully trained autoencoder is not discussed.

\section{Autoencoder Experiments}
Our theoretical analysis of noisy autoencoders shows that NAEs can implement a variety of regularized autoencoder penalties. Here we evaluate the effectiveness of noisy autoencoders at learning representations through a variety of experiments on natural images, MNIST, and CIFAR-10.

All experiments used stochastic gradient descent with momentum to train models. We found that momentum was critical in training both autoencoders and supervised deep networks. Learning rates, batch size, and additional hyperparameters were selected through a combination of manual and automatic grid searches on validation sets. We consider autoencoders with a sigmoidal encoder, linear decoder, and squared error loss. We experimented with isotropic Gaussian input noise with fixed variance $\sigma^2_\mathcal{I}$, isotropic Gaussian hidden unit input noise with fixed variance $\sigma^2_\mathcal{Z}$, and hidden unit activation noise that was either dropout (with inclusion probability $p$), additive zero mean Gaussian with variance $\sigma^2_{\mathcal{H}}$ or multiplicative Gaussian with mean 1 and variance $\sigma^2_{\mathcal{H}}$. Unless otherwise specified, we fix the input noise to be $\sigma^2_{\mathcal{I}}=0.1$, and the number of hidden units to be 1000. All experiments were run in Python using the Pylearn2\footnote{http://github.com/lisa-lab/pylearn2} framework on a single Intel Xeon machine with an NVIDIA GTX 660 GPU.

\subsection{Denoising Natural Images}
In our first experiment, we evaluated the effect of dropout noise on the generalization performance of a noisy autoencoder. We trained two NAEs on 12x12 patches drawn from the van Hateren natural image dataset \cite{vanHateren:1998vn}. The first NAE had noise on the input but not hidden activations (simply a DAE), while the second additionally had dropout noise with $p=0.5$ on the hidden activations. We evaluated denoising performance on an independent set of image patches with noise variance equal to the corrupting input noise of the NAE, and computed the average reconstruction error over 1000 noisy inputs. The NAE with and without dropout had average reconstruction errors of 2.5, and 3.2 respectively. Thus NAEs are able to improve denoising performance over typical DAEs.

\subsection{Effect on MNIST Features}

To better understand the impact of dropout on NAE features, we trained a set of models with varying levels of input noise and dropout noise on hidden activations. We used smaller networks for this experiment, with 250 hidden units and training on only the first 10,000 digits of MNIST.

\begin{figure}[h]
\begin{center}
\includegraphics[width=0.45\textwidth]{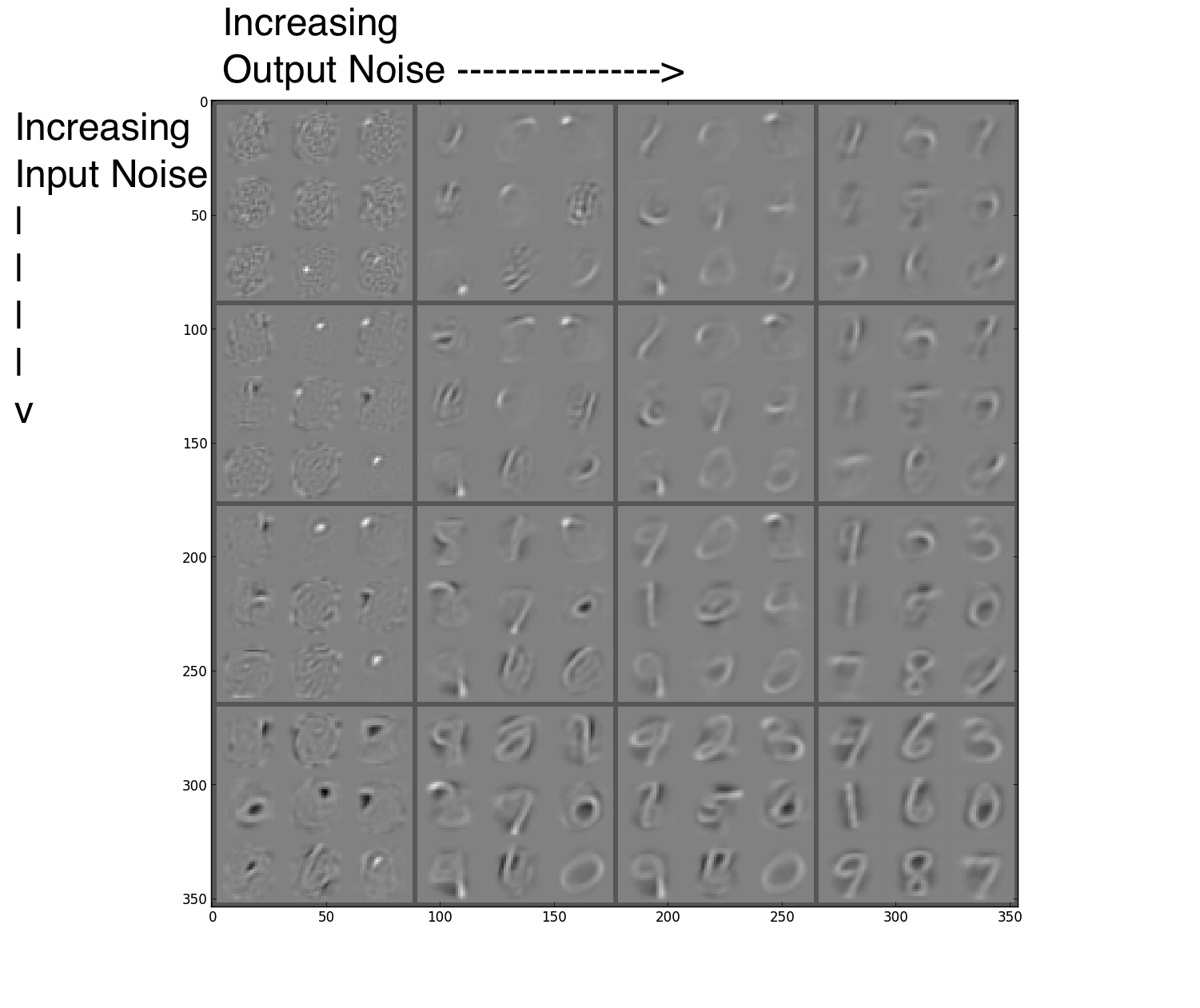}
\includegraphics[width=0.49\textwidth]{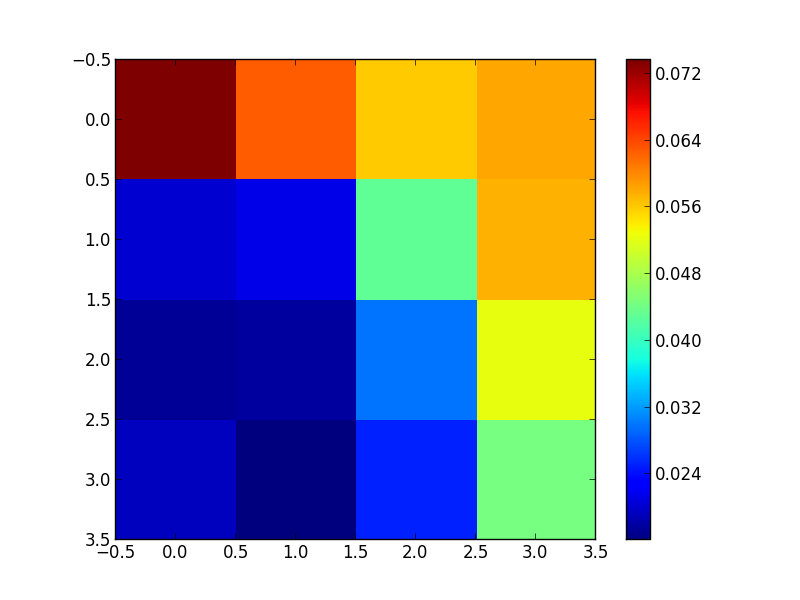}

\end{center}
\caption{Impact of input and hidden activation noise on filters (left) and classification performance (right). Each block in the 4x4 grid corresponds to a different NAE model. The first column is a DAE as there is no output noise.}
\label{fig:feats}

\end{figure}

In Figure \ref{fig:feats} (left), we show the effect of input noise and hidden unit activation noise on the learned features. With no input or hidden activation noise, the noisy autoencoder reduces to a vanilla autoencoder and tends to learn features which do not capture interesting structure in the data. As we increase the input noise, we learn features that capture increasingly larger local structure in the digits, as found in DAEs \cite{Vincent:2010vu}. Increasing hidden activation noise leads to more global features that tend to resemble large strokes or segments of digits. If we increase {\em both} input noise and activation noise, we find features which resemble larger sections of digits. We also evaluated the classification error for these different models by using them to initialize a multilayer perceptron with a softmax classifier on top of the learned hidden representation. Importantly, we find the best performing model for classification contains both input noise and hidden activation noise (Figure \ref{fig:feats}, right).

\subsection{MNIST Classification}
\begin{table}[t]
\caption{Comparison of classification performance for autoencoder models trained on MNIST with 2000 hidden units. Columns indicate different pretraining methods, and rows indicate different supervised training methods}
\label{sample-table}
\begin{center}
\begin{tabular}{|r|c|c|c|c|c|c|c|c|c|c|c|}
\hline
\backslashbox{Fine-Tuning}{Pretraining}
& AE & DAE & CAE & DAE+Poisson& DAE+Dropout &DAE+Gaussian \\
\hline
Backprop Test Errors &164 & 145 & 149 & 143 &  138 &159\\
\hline
Dropout Test Errors&143 & 128 & 109 & 122 & {\bf 99} & 135 \\
\hline
Gaussian Test Errors& 147 & 120 & 115 & 145& 105 & 151\\
\hline
\end{tabular}
\end{center}
\end{table}

\begin{table}[t]
\caption{Comparison of classification performance for autoencoder models trained on MNIST with 2000 hidden units. Columns indicate different pretraining methods, and rows indicate different supervised training methods}
\label{sample-table}
\begin{center}
\begin{tabular}{|r|c|c|c|c|c|c|c|c|c|c|c|}
\hline
&DAE & DAE + Dropout & DAE + Poisson \\
\hline
Accuracy & 71.1\% & 74.6\% & 76.9\%\\
\hline
\end{tabular}
\end{center}
\end{table}

\begin{table}[t]
\caption{Comparison of classification performance for autoencoder models trained on MNIST with 2000 hidden units. Columns indicate different pretraining methods, and rows indicate different supervised training methods}
\label{sample-table}
\begin{center}
\begin{tabular}{|r|c|c|c|c|c|c|c|c|c|c|c|}
\hline
& No noise & Dropout & Gaussian & Poisson \\
\hline
Test Errors & 154 & 110 & 85 & 92\\
\hline
\end{tabular}
\end{center}
\end{table}
To better evaluate the impact of hidden unit input and activation noise on NAE classification performance, we trained larger models with 2000 hidden units and fixed the Gaussian input noise variance at $0.1$. We considered Gaussian noise at the hidden unit inputs, and both dropout and Gaussian noise at the hidden unit activations. These models were used as initialization for a MLP that was trained with standard backpropagation. The level of hidden noise on 10000 heldout examples from the training set was used to optimize the noise level. We also trained standard autoencoders, DAEs, and CAEs of the same architecture. We found that NAEs with a dropout of $p=0.25$ achieved the lowest test error of 138.

Given the pretrained NAE, we can also perform noisy backpropagation where we continue injecting noise into the model while training the classifier. This noise can be the same as when we perform pretraining, but it can also be an entirely different type of noise. We consider performing noisy backpropagation with $\sigma^2_{\mathcal{I}}=0.1$, and hidden noise that is either dropout with $p=0.25$, or Gaussian additive noise with $\sigma^2_{\mathcal{H}}=0.025$. We find that the NAE dropout model tuned with dropout noise achieves a test error of 99. To the best of our knowledge, this is the lowest classification error achieved on MNIST with a single layer model. This model is able to waste less capacity in a large network, but requires regularization while performing supervised fine-tuning to prevent overfitting.

Supervised training with Gaussian hidden noise also improves performance relative to standard backpropagation. However, Gaussian hidden noise does not help very much in the unsupervised features as it simply does weight decay.

\subsection{CIFAR-10 Classification}
To validate the utility of NAEs in other models, we also analyze the CIFAR-10 dataset. We train a NAE with $\sigma^2_\mathcal{I}=0.25$, dropout hidden activation noise with $p=0.5$, and 800 units on 6x6 patches extracted from the CIFAR-10 dataset. Using these features, we extract a high level representation of full size images using the pipeline from \cite{Coates:2010wo}. Training an SVM classifier on the resulting representation yielded an accuracy of 74.5\%. This accuracy is slightly lower than that reported for the higher-order CAE \cite{Rifai:2011uia}, but better than all other regularized autoencoder representations. Training a DAE results in a lower accuracy of 73.6\%, indicating that adding dropout noise while learning features is helpful.

\subsection{Purely Supervised MNIST}
We have shown that different types of noise can be used to regularize hidden representations and improve classification performance on MNIST and CIFAR-10. Furthermore, we found that both dropout and additive Gaussian noise on hidden activations while fine-tuning can improve classification error. Here we experiment with a deep MNIST model from \cite{Hinton:2012tv}. This model consists of two hidden layers of 1200 rectified linear units, and was trained with dropout on the inputs and hidden activations. It was formerly the state-of-the-art result for single models not incorporating prior domain knowledge or pretraining, but has recently been surpassed by maxout networks \cite{Anonymous:5Kr-Pl7w}. Both these networks utilize dropout while training, and the same scaling we perform at test time. Instead of training with dropout, we use other types of noise on the input and hidden unit activations. We experimented with additive Gaussian noise with a fixed variance, and additive Gaussian noise whose variance was proportional to the mean activation (we call this Poisson noise as a shorthand). We optimized these networks using SGD with momentum using the same parameters as in \cite{Hinton:2012tv}. As in previous experiments, we fixed $\sigma^2_\mathcal{I}=0.1$, and selected $\sigma^2_\mathcal{H}$ through cross validation for additive Gaussian noise. We found that our best Gaussian model had 85 errors on the test set and the best Poisson model has 92 errors,  beating both dropout and maxout networks.

\begin{figure}[h]
\begin{center}
\includegraphics[width=0.8\textwidth]{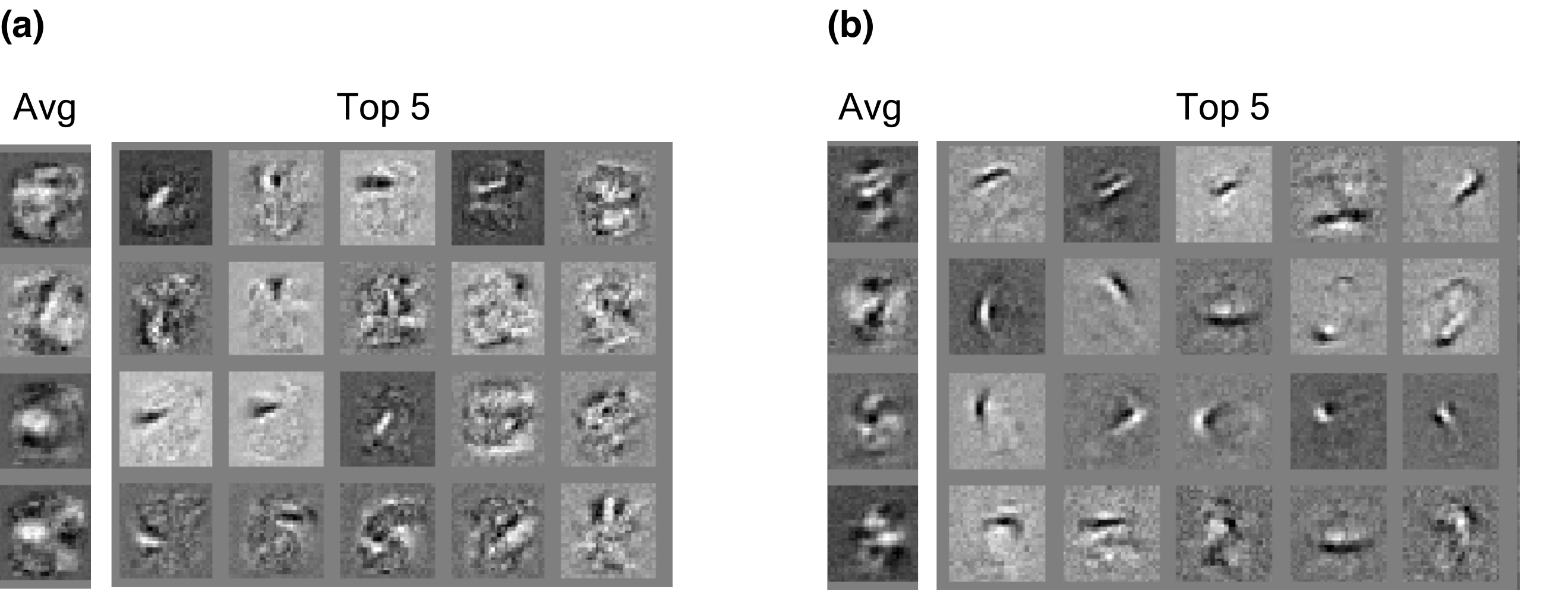}

\end{center}
\caption{Visualization of features learned in the deep MNIST network for (a) dropout and (b) additive Gaussian noise (right). Each row corresponds to a different second layer neuron. The first column indicates the average of all the underlying first layer feature weights times the connection strength to that neuron. The remaining columns show the 5 most strongly connected first layer features }
\label{fig:layer2}
\end{figure}

\begin{figure}[h]
\begin{center}
\includegraphics[width=1.1\textwidth]{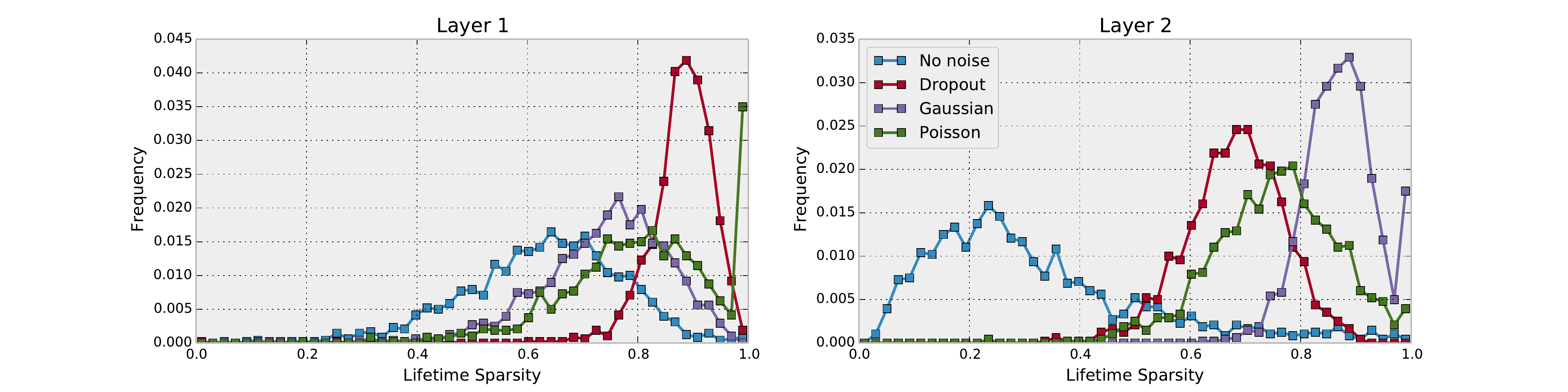}
\includegraphics[width=1.1\textwidth]{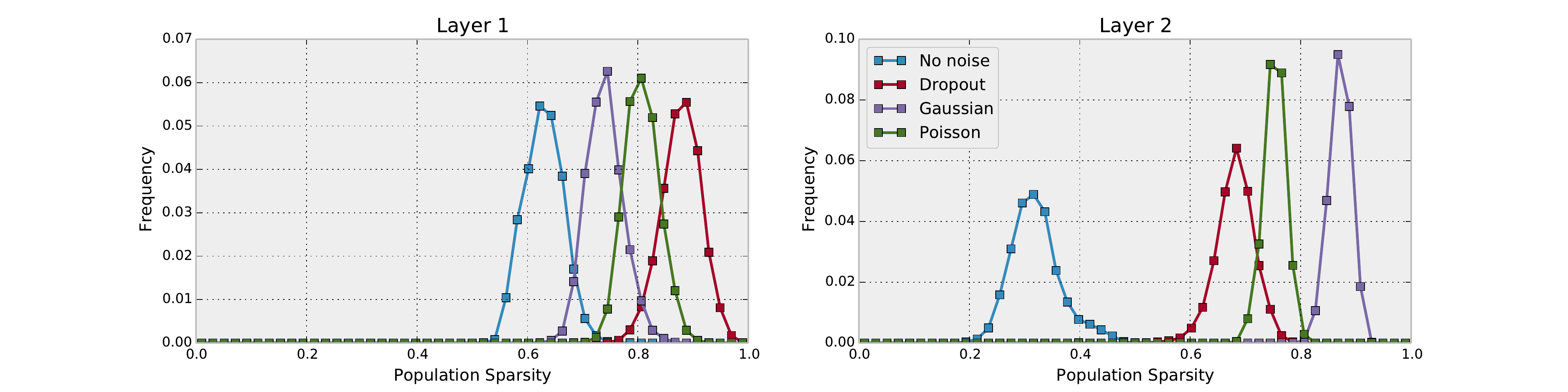}
\includegraphics[width=1.1\textwidth]{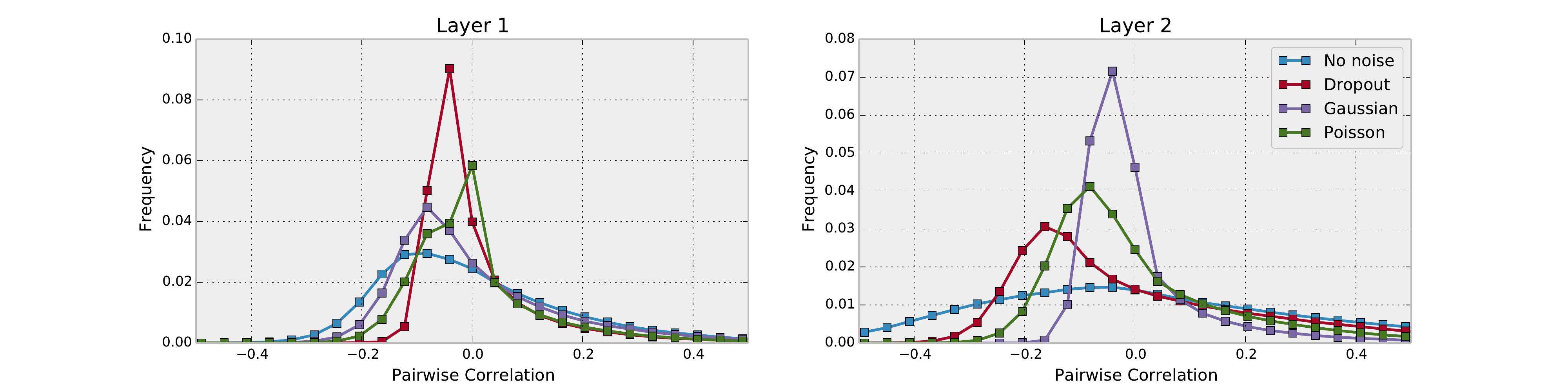}
\includegraphics[width=1.1\textwidth]{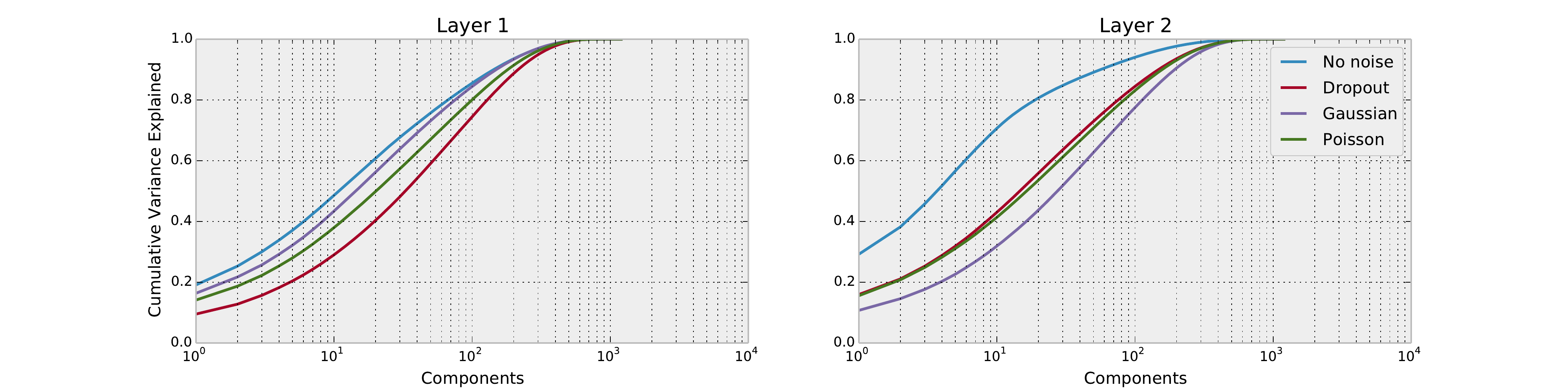}

\end{center}
\caption{Effect of noise type on sparsity and activations at test time. See text for details.}
\label{fig:sparsity}
\end{figure}

To better understand why these other types of noise beats dropout and the noiseless version, we visualized the hidden representation for both networks (Figure \ref{fig:layer2}). We find that Gaussian noise leads to less noisy first layer filters, and tends to group together more similar second layer features. To further understand the influence of noise on the network, we analyzed the activations across the 3 different types of noise, and the noiseless network. We found that all types of noise increase both the lifetime and population sparsity of neurons in both layers of the network. Gaussian noise yielded the sparsest second layer representations, while dropout noise yieldest the sparsest first layer representations both in lifetime and population sparseness (Figure \ref{fig:sparsity}, rows 1 and 2). We also found that noise decorrelated activations relative to the noiseless network (Figure \ref{fig:sparsity}, row 3). This decorrelation with noise also flattened the spectrum of the activations, where the cumulative variance explained grew slower for the noisy networks (Figure \ref{fig:sparsity}, row 4). These results show that  noise acts to sparsity and decorrelate representations, and spread information more evenly across the network.

\section{Discussion}
In summary, by proposing a unifying principle for auto-encoders, namely robustness of auto-encoders to injections of noise at all levels of internal representation, we have generated a framework for viewing a wide variety of previous training algorithms through a unified lens.  Different choices of where to inject noise, and what type of noise, lead to different algorithms.  This allows us to generate many new training algorithms by designing noise, not only in NAE's but also in networks engaged in direct supervised learning.  

Using these techniques we can achieve very good performance on benchmark tasks. We are able to obtain the best performing pretrained one layer model on MNIST, as well as the best performing deep model that does not incorporate prior knowledge or pretraining. However, we stress that we have perhaps not even begun to explore the full power of NAEs, as we have not systematically explored the huge design space for what type of noise to inject, and where and when to inject it.   A great deal of research lies ahead in understanding this design space and generating the algorithms and theories for automatically making data dependent design choices in this space. Furthermore, both the space of noise that is useful for supervised training must also be explored, as well as how noise in NAEs and noise while supervised fine-tuning interact to optimize performance.

Finally, intuitively, why might noise injection in the internal representations of a deep network be a good idea?  We hypothesize that because each training data point has an inherent noise ball around its hidden representation, classification will not be possible unless data points of different classes lie outside each other's noise balls.  This introduces an effective repulsion between data points of different classes in the space of hidden representations. On the other hand, the invariance of the classifier required to properly categorize different inputs of the same class introduces an inherent compressive force in the projective map from the input level to the hidden level.  Thus noise on internal representations leads to a spreading of representations of different classes, and a contraction within classes that may be beneficial for generalization.
\small{
\bibliographystyle{plain}
\bibliography{newlib}}

\begin{thebibliography}{10}

\bibitem{An:1996wk}
Guozhong An.
\newblock {The effects of adding noise during backpropagation training on a
  generalization performance}.
\newblock {\em Neural computation}, 8(3):643--674, 1996.

\bibitem{Bengio:2013ux}
Yoshua Bengio.
\newblock {Deep Learning of Representations: Looking Forward}.
\newblock {\em arXiv.org}, May 2013.

\bibitem{Bishop:1995tf}
Chris~M Bishop.
\newblock {Training with noise is equivalent to Tikhonov regularization}.
\newblock {\em Neural computation}, 7(1):108--116, 1995.

\bibitem{Chen:2012tq}
Minmin Chen, Zhixiang Xu, Kilian Weinberger, and Fei Sha.
\newblock {Marginalized Denoising Autoencoders for Domain Adaptation}.
\newblock {\em arXiv.org}, June 2012.

\bibitem{Cho:2013wc}
Kyunghyun Cho.
\newblock {Boltzmann Machines and Denoising Autoencoders for Image Denoising}.
\newblock January 2013.

\bibitem{Coates:2010wo}
A~Coates, H~Lee, and A~Y Ng.
\newblock {An analysis of single-layer networks in unsupervised feature
  learning}.
\newblock {\em Ann Arbor}, 1001:48109, 2010.

\bibitem{Doi:2012vi}
E~Doi, J~L Gauthier, G~D Field, and J~Shlens.
\newblock {Efficient coding of spatial information in the primate retina}.
\newblock {\em The Journal of {\ldots}}, 2012.

\bibitem{Goodfellow:2009vda}
Ian Goodfellow, Quoc Le, Andrew Saxe, Honglak Lee, and Andrew~Y Ng.
\newblock {Measuring invariances in deep networks}.
\newblock {\em Advances in neural information processing systems}, 22:646--654,
  2009.

\bibitem{Anonymous:5Kr-Pl7w}
Ian~J Goodfellow, David Warde-Farley, Mehdi Mirza, Aaron Courville, and Yoshua
  Bengio.
\newblock {Maxout Networks}.
\newblock {\em arXiv.org}, February 2013.

\bibitem{Hinton:2012tv}
Geoffrey~E Hinton, Nitish Srivastava, Alex Krizhevsky, Ilya Sutskever, and
  Ruslan~R Salakhutdinov.
\newblock {Improving neural networks by preventing co-adaptation of feature
  detectors}.
\newblock {\em arXiv.org}, July 2012.

\bibitem{Inayoshi:2005iz}
H~Inayoshi and T~Kurita.
\newblock {Improved Generalization by Adding both Auto-Association and
  Hidden-Layer-Noise to Neural-Network-Based-Classifiers}.
\newblock {\em Machine Learning for Signal Processing, 2005 IEEE Workshop on},
  pages 141--146, 2005.

\bibitem{Karklin:N4TxuNDQ}
Yan Karklin and Eero Simoncelli.
\newblock {Efficient coding of natural images with a population of noisy
  Linear-Nonlinear neurons}.

\bibitem{Anonymous:H68d5mS5}
Alex Krizhevsky, Ilya Sutskever, and Geoff Hinton.
\newblock {Imagenet classification with deep convolutional neural networks}.
\newblock pages 1106--1114, 2012.

\bibitem{Anonymous:DmtqWbpa}
Quoc~V Le, Alexandre Karpenko, Jiquan Ngiam, and Andrew~Y Ng.
\newblock {Ica with reconstruction cost for efficient overcomplete feature
  learning}.
\newblock {\em Advances in neural information processing systems},
  24:1017--1025, 2011.

\bibitem{Lee:2009tm}
Honglak Lee, Roger Grosse, Rajesh Ranganath, and Andrew~Y Ng.
\newblock {Convolutional deep belief networks for scalable unsupervised
  learning of hierarchical representations}.
\newblock pages 609--616, 2009.

\bibitem{Anonymous:g1ChF3mx}
S~Rifai, P~Vincent, X~Muller, X~Glorot, and Y~Bengio.
\newblock {Contractive auto-encoders: Explicit invariance during feature
  extraction}.
\newblock 2011.

\bibitem{Rifai:2011td}
Salah Rifai, Xavier Glorot, Yoshua Bengio, and Pascal Vincent.
\newblock {Adding noise to the input of a model trained with a regularized
  objective}.
\newblock {\em arXiv.org}, April 2011.

\bibitem{Rifai:2011uia}
Salah Rifai, Gr{\'e}goire Mesnil, Pascal Vincent, Xavier Muller, Yoshua Bengio,
  Yann Dauphin, and Xavier Glorot.
\newblock {Higher order contractive auto-encoder}.
\newblock In {\em ECML PKDD'11: Proceedings of the 2011 European conference on
  Machine learning and knowledge discovery in databases}. Springer-Verlag,
  September 2011.

\bibitem{Salakhutdinov:2007wy}
Ruslan Salakhutdinov and Geoffrey Hinton.
\newblock {Semantic hashing}.
\newblock {\em RBM}, 500(3):500, 2007.

\bibitem{vanHateren:1998vn}
J~Hans van Hateren and Arjen van~der Schaaf.
\newblock {Independent component filters of natural images compared with simple
  cells in primary visual cortex}.
\newblock {\em Proceedings of the Royal Society of London. Series B: Biological
  Sciences}, 265(1394):359--366, 1998.

\bibitem{Vincent:2010vu}
P~Vincent, H~Larochelle, I~Lajoie, Y~Bengio, and P~A Manzagol.
\newblock {Stacked denoising autoencoders: Learning useful representations in a
  deep network with a local denoising criterion}.
\newblock {\em Journal of Machine Learning Research}, 11:3371--3408, 2010.

\bibitem{Wang:0xFoHXs6}
Sida~I Wang and Christopher~D Manning.
\newblock {Fast dropout training}.

\bibitem{Zeiler:2013tna}
Matthew~D Zeiler and Rob Fergus.
\newblock {Stochastic Pooling for Regularization of Deep Convolutional Neural
  Networks}.
\newblock {\em arXiv.org}, January 2013.

\end{thebibliography}

\end{document}